\documentclass{article}
     \PassOptionsToPackage{square,sort,comma,numbers}{natbib}

    \usepackage[preprint]{neurips_2020}

\usepackage[utf8]{inputenc} 
\usepackage[T1]{fontenc}    
\usepackage{hyperref}       
\usepackage{url}            
\usepackage{booktabs}       
\usepackage{amsfonts}       
\usepackage{nicefrac}       
\usepackage{microtype}      
\usepackage{color,colortbl}
\usepackage{mwe}
\usepackage{caption}
\usepackage{subcaption} 
\usepackage{appendix}
\usepackage{dirtytalk}
\usepackage{amsfonts}
\usepackage{graphicx}
\usepackage{amsmath}
\usepackage{mathtools}
\usepackage{siunitx}

\newcommand{\eg}{e.g.\xspace}
\newcommand{\ie}{i.e.\xspace}

\newcommand{\wrt}{with respect to\xspace}

\newcommand{\niid}{non-i.i.d.\xspace}
\newcommand{\ts}{\textsuperscript}

\DeclareMathOperator*{\argmin}{arg\,min}

\DeclareMathOperator{\sign}{sgn}

\title{Improving Accuracy of Federated Learning \\ in Non-IID Settings}

\author{%
  Mustafa Safa Ozdayi \\
  The University of Texas at Dallas \\
  \texttt{mustafa.ozdayi@utdallas.edu} \\
  
  \And
  Murat Kantarcioglu \\
  The University of Texas at Dallas \\
  \texttt{muratk@utdallas.edu} \\
  
  \And
  Rishabh Iyer \\
  The University of Texas at Dallas \\
  \texttt{rishabh.iyer@utdallas.edu} \\
  
}

\begin{document}
\maketitle
\begin{abstract}
 Federated Learning (FL) is a decentralized machine learning protocol that allows a set of participating agents to collaboratively train a model without sharing their data.
 This makes FL particularly suitable for settings where data privacy is desired.
 However, it has been observed that the performance of FL is closely tied with the local data distributions of agents. Particularly, in settings where local data distributions vastly differ among agents, FL performs rather poorly \wrt the centralized training. To address this problem, we hypothesize the reasons behind the performance degradation, and develop some techniques to address these reasons accordingly. In this work, we identify four simple techniques that can improve the performance of trained models without incurring any additional communication overhead to FL, but rather, some light computation overhead either on the client, or the server-side. In our experimental analysis,  a combination of our techniques improved the validation accuracy of a model trained via FL by more than 12\% \wrt our baseline. This is about 5\% less than the accuracy of the model trained on centralized data.

\end{abstract}

\section{Introduction}
Federated Learning (FL)~\cite{fed-learning:google} is a distributed, and decentralized protocol to train machine learning models. A set of participating agents can jointly train a model without sharing their local data with each other, or any other third-party. In that regard, FL differs from the traditional distributed learning setting in which data is first centralized, and then distributed to the agents~\cite{dean2012large,li2014scaling}. 
 Due to this, FL is expected to have prominent applications in settings where data privacy is of concern.
  
 
However, some recent works have observed that the accuracy of models drops significantly in FL as the local data distributions of agents differ~\cite{zhao2018federated:noniid, hsieh2019non:noniid}. In this work, we study this phenomena, and identify a few techniques that improves performance of FL when data is distributed in a \niid fashion among the participating agents. In contrast to some of the recent works, the techniques we identify incur no extra communication overhead to the FL, \eg, server does not have to transmit any data to agents as in~\cite{zhao2018federated:noniid}, or the trained model does not have to be shuffled across participating agents as in~\cite{hsieh2019non:noniid}. In brief, techniques we identify improve the performance of FL only by incurring some computation overhead either on the client-side, or the server-side. Further, they can easily be incorporated to FL without any structural changes.

We organize the rest of the paper as follows. In Section~\ref{sec:background}, we provide the necessary background on FL. In Section~\ref{sec:methods}, we discuss, and explain our techniques to improve FL in \niid setting. In Section~\ref{sec:experiments}, we provide the experimental results, and demonstrate the performance improvement in FL due to our techniques. Finally in Section~\ref{sec:conclusion}, we provide a few concluding remarks.
\section{Background}\label{sec:background}
\subsection{Federated Learning (FL)}
At a high level, FL  is multi-round protocol between an aggregation serve,r and a set of agents where agents jointly train a machine learning model. Formally, participating agents try to minimize the average of their loss functions,
$$
\argmin_{w \in R^d} f(w) = \frac{1}{K}\sum_{k=1}^K f_k(w),
$$
where $f_k$ is the loss function of k\ts{th} agent. For example, for neural networks, $f_k$ is typically empirical risk minimization under a loss function $L$, \ie,

$$
f_k(w) = \frac{1}{n_k} \sum_{i=1}^{n_k} L(w, x_i),
$$

with $n_k$ being the total number of samples in agent's dataset and $x_i$ being the i\ts{th} sample.

A round $r$ in FL starts by the serving picking a subset of agents $S_r$, and sending them the current weights of the model $w_r$. After receiving the weights, an agent initializes his model with $w_r$, and trains for some number of iterations on his local dataset, for example via stochastic gradient descent (SGD). At the end of his local training, i\ts{th} agent computes his update for the current round as $\Delta_r^i = w_r^i - w_r$, and sends the update back to the server. Once server receives updates from all the agents in $S_r$, it computes the weights for the next round via weighted averaging\footnote{Technically, the server can use an arbitrary aggregation function, but we stick to weighted averaging in this work as it is the most common one.} as,

\begin{equation}
w_{r+1} = w_r + \eta \frac{\sum_{k \in S_r} n_k \cdot \Delta_r^k}{\sum_{k \in S_r} n_k}
\label{eqn:fedavg}
\end{equation}
where $n_k$ is the number of samples in the dataset of agent $k$. Aggregating the updates via this way has been referred as FedAvg~\cite{fed-learning:google}.

It has been shown that models trained via FL can perform better than locally trained models at agent's side in various settings ~\cite{fed-learning:google, keyboard}. In contrast, as noted before, it has also been observed that the performance of FL drops drastically when local data distributions of agents differ significantly, \ie, when data is distiributed in a \niid fashion among agents~\cite{zhao2018federated:noniid, hsieh2019non:noniid}.

\section{Methods}
\label{sec:methods}
In this section, we discuss a few possible causes for performance degradation of FL in \niid setting, and propose techniques to alleviate these causes.

\subsection{Hypothesis Conflicts and Server-Side Training}\label{sec:pred_conf}
\begin{figure}[h]
  \begin{subfigure}[b]{\textwidth}
  \center
    \includegraphics[scale=0.35]{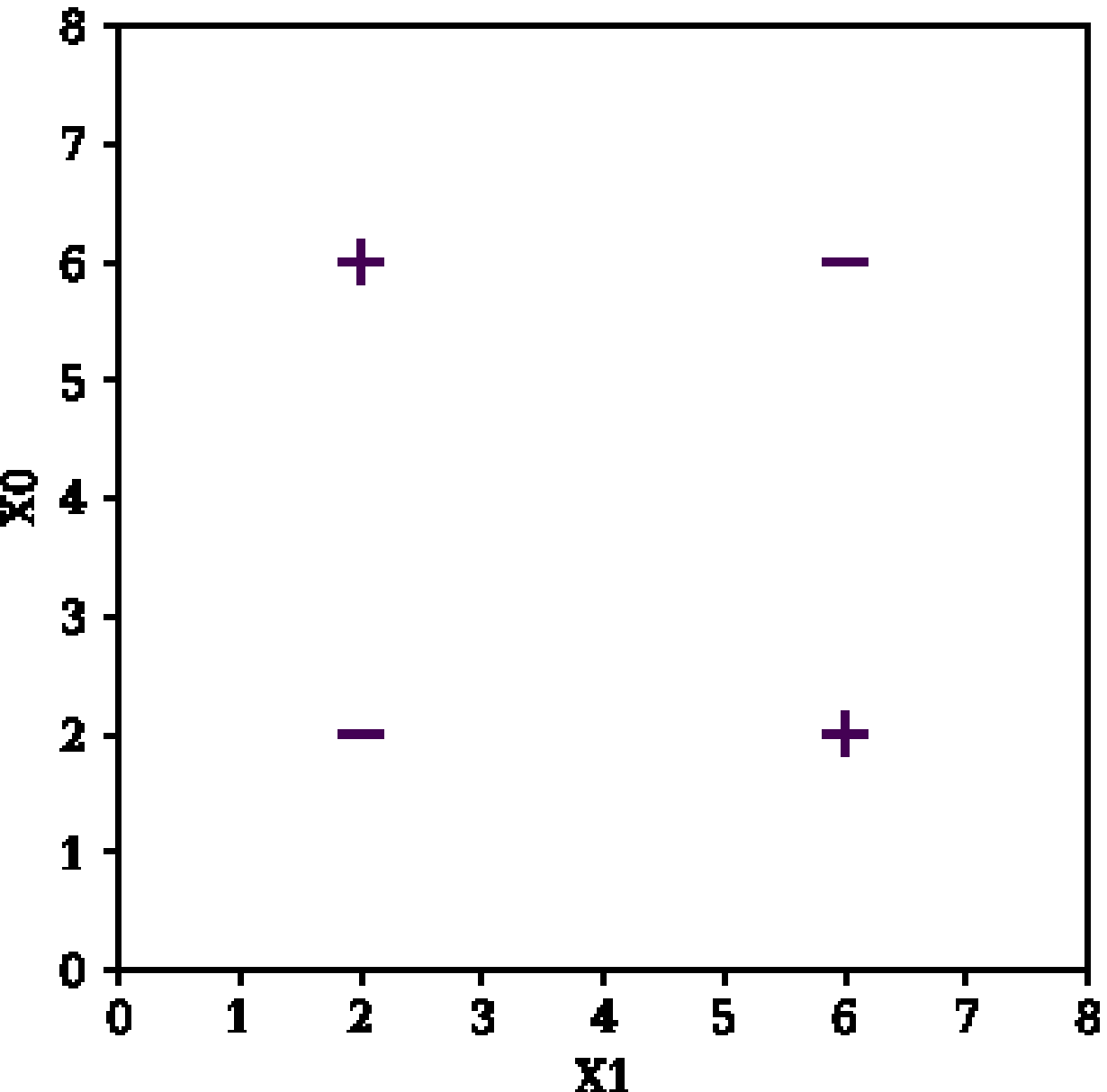}
    \includegraphics[scale=0.35]{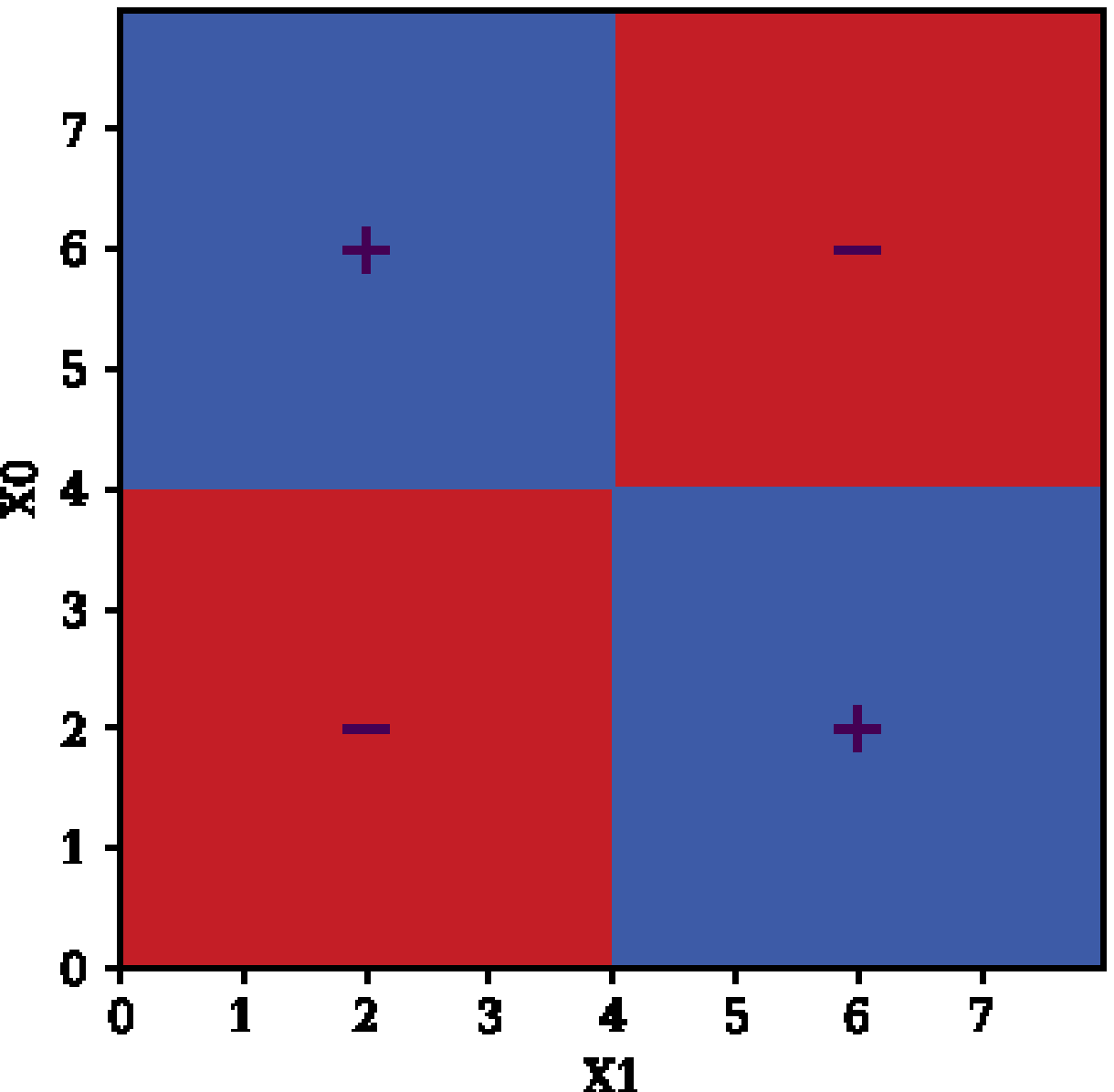}
   \caption{Prediction in centralized setting}
  \end{subfigure}
  \vfill
  \begin{subfigure}[b]{\textwidth}
  \center
   \includegraphics[scale=0.35]{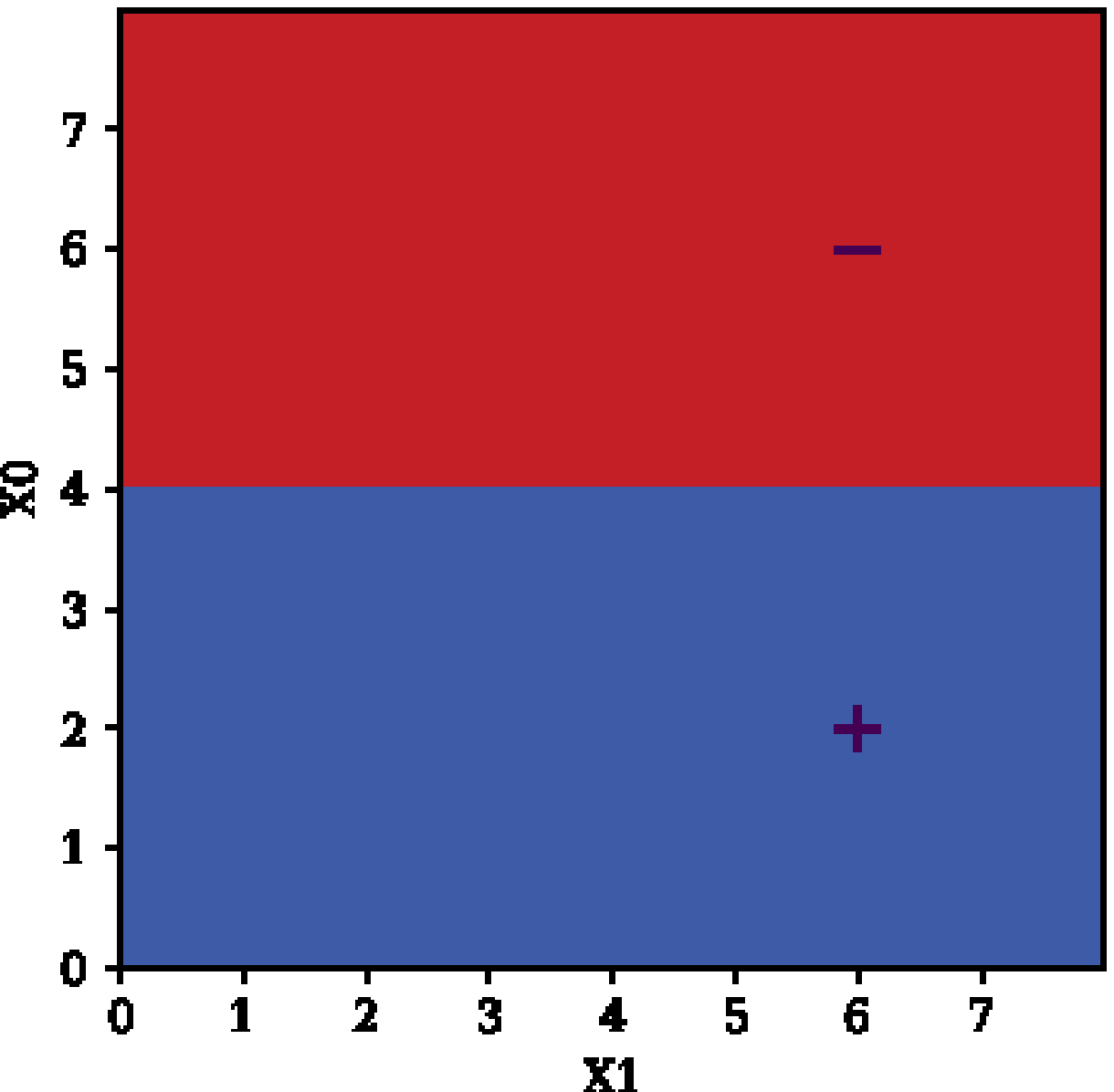}
   \includegraphics[scale=0.35]{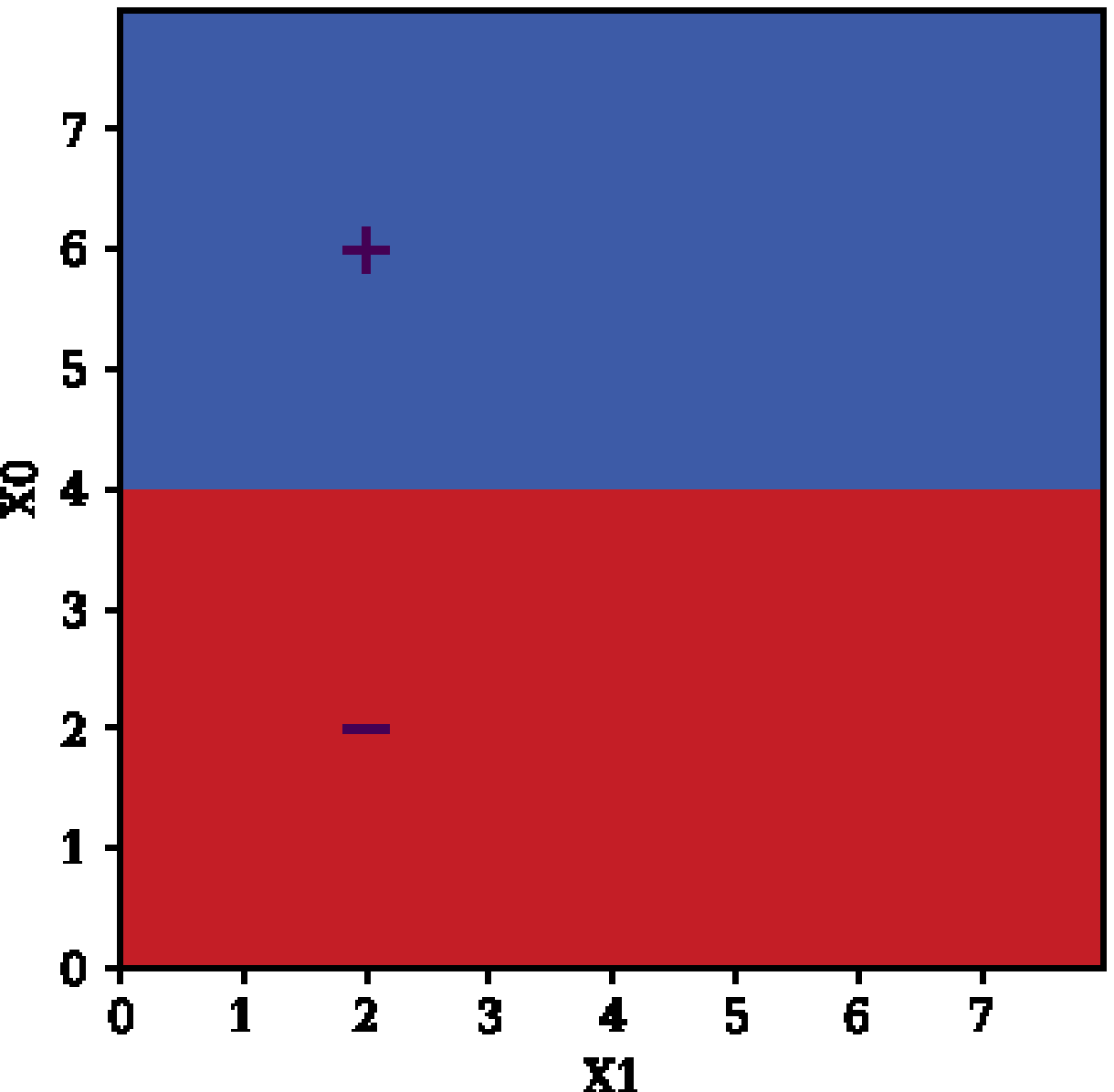}
  \caption{Hypothesis conflict between two sites in distributed setting}
  \end{subfigure}
  \caption{An illustration of how hypothesis conflicts can lead to performance degradation. In (a), we have a simple dataset of four points with two features. As can be seen, a decision tree can easily learn the rules to classify this dataset perfectly. In (b), the dataset is partitioned where first agent gets points with $x_1 > 4$ (left), and the second agent gets points with $x_1 < 4$ (right). As can be seen, agents' local models learn conflicting hypotheses. First agent predicts (-) if $x_0 > 4$ and second agent predicts (+) if $x_0 > 4$. If we were to do ensemble prediction with these models, their votes would just cancel each other.}
  \label{fig:pred_conflicts}
\end{figure}

\paragraph{Hypothesis conflicts}
We hypothesize that one cause behind the performance degradation in \niid setting is \emph{hypothesis conflicts} between agents' local models. For example, for a classification task, we say two models have conflicting hypotheses if there exists an input in which the models assign different classes. Concretely, let $f_{i, r}, f_{j, r}$ be the local models of agent $i, j$ at the end of round $r$, respectively. Then, we say there is a hypothesis conflict between these models at round $r$ if there exists an input $x$ such that $f_{i, r}(x) \neq f_{j, r}(x)$.
We illustrate how hypothesis conflicts can be problematic on a toy example in Figure~\ref{fig:pred_conflicts}.

\paragraph{Server-side training}
Given the discussion above, we argue that having a small, balanced training data on server-side can improve the performance in \niid setting significantly. That is, the central server can aggregate the updates as usual, and then fine-tune the resulting model on such a data by training the model for a short amount.
If the training data on the server-side has a similar distribution \wrt the union of the datasets of agents, then the parameters learned on this data will likely perform well on the local data of agents. Given that, agents should not overwrite the mappings learned during server-side training. We believe this can reduce the hypothesis conflicts, and consequently improve the performance of the trained models.

\subsection{Projected Gradient Descent at Agents' Side}
\begin{figure}[!ht]
    \centering
    \includegraphics[scale=0.65]{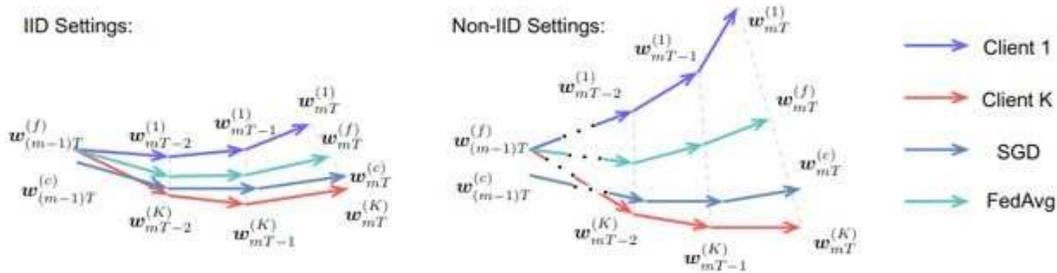}
    \caption{Parameter divergence in FL \cite{zhao2018federated:noniid}. SGD refers to the direction of the gradient in centralized setting. As local loss surfaces differ, local agents' parameters diverge from each other. This could be one of the reasons why FL performs significantly worse in \niid setting, and perhaps can be alleviated to some extend by using projected gradient descent,}
    \label{fig:paramdiv}
\end{figure}

Another reason for the performance degradation might stem from the fact that local loss surfaces of agents are different. Each local model might reach to some local minima on their own surface, however, aggregated model might not be close to any local minima on the loss surface defined by the union of local datasets. Indeed, in a recent work \cite{zhao2018federated:noniid}, authors hypothesize FL performs worse in \niid setting as local models' parameters differ significantly from each other (see Figure \ref{fig:paramdiv}).

We believe that this divergence can be alleviated to some extent by having each agent run projected gradient descent with some common parameters, e.g., $L_2$ norm of local models can be bounded by a common value $M$. Since each local model is initialized to the same parameters, and each agent uses the same projection, we believe this might prevent local models from diverging too much from each other. Consequently, this can improve the performance of the trained models.

\subsection{Server-side Momentum}
It is well known that momentum techniques can significantly improve convergence by making gradient descent robust to certain elements in loss surfaces, e.g., it can dampen oscillations in \say{valleys}. Given the differences in loss surfaces, we believe that it might have a similar effect in \niid case. To adapt momentum in FL, we keep running averages of aggregated models. That is, with momentum, the update rule for FedAvg (equation \ref{eqn:fedavg}) can be written as,

\begin{equation}
v_{r+1} = \beta v_r + \frac{\sum_{k \in S_r} n_k \cdot \Delta_r^k}{\sum_{k \in S_r} n_k}
\label{eqn:momentum}
\end{equation}

\begin{equation}
w_{r+1} = v_{r+1} + \eta w_r
\label{eqn:momentum}
\end{equation}

where $\beta$ is server's momentum constant and $v_0 = 0$.

\subsection{Preventing Parameter Cancellation}
\label{sec:lr_idea}
Following the argument in Figure~\ref{fig:paramdiv}, we predict that as local models' diverge, they could cancel each others' parameters when they are aggregated. To prevent this, we can simply look at the sign information of updates, and adjust the learning rate $\eta$ at the server. That is, we can set a threshold $\theta$, and for every dimension where the sum of signs of updates is less than this threshold, we can set $\eta$ to 0. With this, the server's learning rate for i\ts{th} dimension is given by,
$$
\eta_{\theta, i} =
\begin{cases}
0 & \sum_{k \in S_t} \sign(\Delta_t^k)_i < \theta \\
\eta  & \text{otherwise}
\end{cases}
$$
With such a learning rate, FedAvg can be written as,
\begin{equation}
w_{r+1} = w_r + 
\eta_\theta \odot
\frac{\sum_{k \in S_r} n_k \cdot \Delta_r^k}{\sum_{k \in S_r} n_k}.
\label{eqn:fedavg_robust}
\end{equation}
where $n_{\theta}$ is simply the learning rate vector over all dimensions, and $\odot$ is the element-wise product operation.

We note that, the same technique has been proposed as a defense against backdoor attacks~\cite{safa2020defending} in FL setting. Although authors note their technique can potentially improve performance of models in \niid setting too, they provide no empirical evaluation.

\section{Experiments}\label{sec:experiments}
In this section, we illustrate the performance of our techniques from Section \ref{sec:methods} via experiments. We first look at each technique in isolation, and then in combination with each other.

\subsection{Setting}
In our experiments, we simulate FL for $R = 100$ rounds among $K = 2$ agents. At each round, agents locally train for $E=1$ epoch with a batch size of $B=256$, using a learning rate of $0.1$, and a weight decay of $\num{5e-4}$ before sending their updates. Upon receiving and aggregating updates, we  measure validation accuracy on a validation data. We use the FedAvg aggregation (equation \ref{eqn:fedavg}) with $\eta=1$. For dataset, we use CIFAR10 ~\cite{cifar10} and for model, we use ResNet20 with fixup initialization ~\cite{zhang2019fixup}.
Each experiment is repeated for 3 times, and we report on the mean and standard deviation  of results.

\subsection{Results}
We first provide the results for baseline in Table \ref{tab:baseline}.
\begin{table}[h!]
\centering
    \begin{tabular}{c|c|c}
   Setting & Val. Acc. Mean & Val. Acc. STD \\
    \hline
    Centralized & 90.4\% & 0.77\% \\
    FL - IID & 90.0\% & 0.08\% \\
    FL - NIID(5) & 73.0\% & 2.6\% \\
    \end{tabular}
\caption{Results of baseline setting. As can be seen, when data is distributed in iid fashion among agents (FL - IID), i.e, each agent gets same number of instances from all 10 classes, accuracy of FL is more or less the same as centralized setting. However, when data is distributed in a way such that first agent gets all samples of first 5 classes, and the other agent gets the remaining samples (FL - NIID(5)), accuracy drops by more than 17\% \wrt the centralized case. }
\label{tab:baseline}
\end{table}

We then illustrate how server-side training, as described in Section \ref{sec:methods}.2, improves upon baseline in Table \ref{tab:server-side}.

\begin{table}[h!]
\centering
    \begin{tabular}{c|c|c|c}
   Setting & Server-side data & Val. Acc. Mean & Val. Acc. STD \\
    \hline
    FL - NIID(5) & 0\% & 73.0\% & 2.6\% \\
    FL - NIID(5) & 5\% & 83.7\% & 0.1\% \\
    \end{tabular}
\caption{Effect of server-side training. We give 5\% of training data to the server and distribute the remaining training data to the agents in \niid fashion as explained in baseline setting. Server fine-tunes the aggregated model by training for a single epoch on this data at each round after aggregating the updates. We see that this alone improves the accuracy over baseline by more than 10\%.}
\label{tab:server-side}
\end{table}

We now show the results for projected gradient descent. In this setting, we use a $L_2$ norm bound $M$ on local models. That is, if at any point $L_2$ norm of his model exceeds $M$, agent projects the model back to the $L_2$ ball with radius $M$. We have also combined this with noise addition as some works argue that noise addition improves regularization \cite{noh2017regularizing}, i.e., at each iteration, agent samples a Gaussian noise with mean $0$ and some standard deviation, and adds it to his gradients.

\begin{table}[!h]
\centering
    \begin{tabular}{c|c|c|c|c}
   Setting & $L_2$ Norm Threshold & Noise STD & Val. Acc. Mean & Val. Acc. STD \\
    \hline
    FL - NIID(5) & 0 & 0 & 73.0\% & 2.6\% \\
    FL - NIID(5) & 3 & 0 & 77.5\% & 2.3\% \\
    FL - NIID(5) & 3 & \num{1e-4} & 79.6\% & 1.3\% \\
    \end{tabular}
\caption{Improvement due to projected gradient descent. As can be seen, using projected gradient descent alone improves over baseline by about 4\%. Combining it with noise addition improves accuracy by more than 6\% \wrt baseline.}
\label{tab:projectedGD}
\end{table}

We now illustrate the effect of server-side momentum in Table \ref{tab:servmoment}.
\begin{table}[h!]
\centering
    \begin{tabular}{c|c|c|c|c}
   Setting & Server's Momentum Constant & Val. Acc. Mean & Val. Acc. STD \\
    \hline
    FL - NIID(5) & 0 & 73.0\% & 2.6\% \\
    FL - NIID(5) & 0.5 & 80.9\% & 2.4\% \\
    FL - NIID(5) & 0.9 & 75.9\% & 1.1\% \\
    \end{tabular}
\caption{Effect of using momentum on server-side. As can be seen, using momentum with a momentum constant of $0.5$ alone improves accuracy over baseline by more than 7\%.}
\label{tab:servmoment}
\end{table}

Finally, we illustrate the effect of adjusting learning rate of server, $\eta$, as described in Section~\ref{sec:lr_idea} in Table \ref{tab:robustLR}. Since we have only two agents, we use a threshold of $\theta = 2$.

\begin{table}[h!]
\centering
    \begin{tabular}{c|c|c|c}
   Setting & Using Adjustable LR & Val. Acc. Mean & Val. Acc. STD \\
    \hline
    FL - NIID(5) & No & 73.0\% & 2.6\% \\
    FL - NIID(5) & Yes & 78.5\% & 1.72\% \\
    
    \end{tabular}
\caption{Effect of adjusting the learning rate at the server-side. The validation accuracy is improved by more than 5\% \wrt the baseline.}
\label{tab:robustLR}
\end{table}

After looking at each method individually, we do a grid search to see if a combination of them can yield a better performance. The result of grid search yielded a mean accuracy of $85.7\%$ with a standard deviation of $0.3\%$ under the following setting: server-side training with 5\% of data, server-side momentum with a constant of $0.9$, and  with adjustable learning rate at the server-side. This is more than a 12\% improvement over our baseline setting.
\section{Conclusion}\label{sec:conclusion}
In this paper, we studied FL with a particular focus on improving the accuracy of trained models in \niid settings. Through our study, we have identified a few simple techniques that improves over our baseline accuracy by more than 12\%. The techniques we identify are rather simple, and can easily be incorporated to FL without any structural changes. Further, they incur no extra communication overhead, but only some light computation overhead either on the client-side, or the server side. In a future work, we hope to expand our experimental setting by testing different models, datasets, and hyperparameters.

\newpage
\bibliographystyle{unsrt}
\bibliography{ml}

\end{document}